\documentclass[11pt,a4paper]{article}
\usepackage[hyperref]{naaclhlt2018}
\usepackage{times}
\usepackage{latexsym}
\usepackage{graphicx}  
\usepackage{url}
\usepackage{amssymb}
\usepackage{subfigure}

\aclfinalcopy 

\title{Syntactic Recurrent Neural Network for Authorship Attribution }
\author{Fereshteh Jafariakinabad, Sansiri Tarnpradab, Kien A. Hua\\
 University of Central Florida\\
  Computer Science Department\\
{\tt fereshteh.jafari@knights.ucf.edu}
}
\date{}

\begin{document}
\maketitle

\begin{abstract}
Writing style is a combination of consistent decisions at different levels of language production including lexical, syntactic, and structural associated to a specific author (or author groups). While lexical-based models have been widely explored in style-based text classification, relying on content makes the model less scalable when dealing with heterogeneous data comprised of various topics. On the other hand, syntactic models which are content-independent, are more robust against topic variance. In this paper, we introduce a syntactic recurrent neural network to encode the syntactic patterns of a document in a hierarchical structure. The model first learns the syntactic representation of sentences from the sequence of part-of-speech tags. For this purpose, we exploit both convolutional filters and long short-term memories to investigate the short-term and long-term dependencies of part-of-speech tags in the sentences. Subsequently, the syntactic representations of sentences are aggregated into document representation using recurrent neural networks. Our experimental results on PAN 2012 dataset for authorship attribution task shows that syntactic recurrent neural network outperforms the lexical model with the identical architecture by approximately 14\% in terms of accuracy.
\end{abstract}

\section{Introduction}
Individuals express their thoughts in different ways due to many factors including, the conventions of language, educational background, and intended audience, etc. In written language, the combination of consistent conscious or unconscious decisions in language production, known as writing style, has been studied widely. Early work on computational stylometry was introduced in the 1960s by Mosteller and Wallace on federalist papers \cite{mosteller1964inference}. Unprecedented availability of digital data in recent years along with the advancements in machine learning techniques has led to an increase in scholarly attention to the field of Computational stylometry \cite{koppel2009computational,neal2017surveying}.


Stylistic features are generally \textit{content-independent} which means that they are mainly consistent across different documents written by a specific author or author groups. Lexical, syntactic, and structural features are three main families of stylistic features. Lexical features represent author's character and word use preferences, while syntactic features capture the syntactic patterns of sentences in a document. Structural features reveal information about how an author organizes the structure of a document.
 
One of the basic problems which is rarely addressed in the literature is the interaction of style and content. While content words can be predictive features of authorial writing style due to the fact
that they carry information about author’s lexical choice, excluding content words as features is a fundamental step for avoiding topic detection rather than style detection \cite{argamon1998style}. However, syntactic and structural features are content-independent which makes them robust against divergence of topics.


The early proposed methods in style detection are conventional machine learning techniques which are based on count-based features. Deep neural networks, although have been widely explored later on in several domains of natural language processing, only few studies have employed this approach to stylometry and authorship attribution \cite{gkagalaauthorship}. The adopted approaches in deep neural network for style-based text classification mainly focus on lexical features despite the fact that lexical-based language models have very limited scalability when dealing with dataset containing diverse topics and genre.

\newpage
While previously proposed deep neural network approaches focus on lexical level, we introduce a syntactic recurrent neural network which hierarchically learns and encodes the syntactic structure of documents. First, the syntactic representation of sentences are learned from the sequence of part-of-speech (POS) tags and then they aggregate into document representation using recurrent neural networks. Afterwards, we use attention mechanism to reward the sentences which contribute more to the detection of authorial writing style. In order to investigate the effect of long-term and short-term dependencies of POS tags in a sentence, we employ long short-term memory (LSTM) and convolutional neural networks (CNN) respectively. The proposed model is expected to be more effective than the conventional count-based models.

The remainder of this paper is organized as follows. In Section \ref{related_work}, we review the proposed methods in the literature for style-based text classification. We elaborate our proposed approach in Section \ref{proposed_method}. In Section \ref{experiment_eval}, we discuss the dataset followed by performance study. Finally, we conclude the paper in Section \ref{conclusions}.

\section{Related Work}\label{related_work}

Writing style is a combination of consistent decisions at different levels of language production including lexical, syntactic, and structural associated to a specific author (or author groups, e.g. female authors or teenage authors) \cite{daelemans2013explanation}. Nowadays, computational stylometry has a wide range of applications in literary science \cite{kabbara2016stylistic, van2017exploring}, forensics \cite{brennan2012adversarial,afroz2012detecting,wang2017liar}, and psycholinguistics \cite{newman2003lying,pennebaker1999linguistic}. Style-based text classification was proposed by Argamon-Engelson et al. \cite{argamon1998style}. The authors used basic stylistic features (the frequency of function words and part-of-speech trigrams) to classify news documents based on the corresponding publisher (newspaper or magazine) as well as text genre (editorial or news item). 

\subsection{Syntax for Style Detection}
Syntactic n-grams are shown to achieve promising results in different stylometric tasks including author profiling task \cite{posadas2015syntactic} and author verification task \cite{krause2014behavioral}.
In particular, Raghavan et al. investigated the use of syntactic information by proposing a probabilistic context-free grammar for the authorship attribution purpose, and used it as a language model for classification \cite{raghavan2010authorship}. 
A combination of lexical and syntactic features has also shown to enhance the model performance.
Sundararajan et al. argue that, although syntax can be helpful for cross-genre authorship attribution, combining syntax and lexical information can further boost the performance for cross-topic attribution and single-domain attribution \cite{sundararajan2018represents}.
Further studies which combine lexical and syntactic features include \cite{soler2017relevance, schwartz2017effect, kreutz2018exploring}

\subsection{Neural Network in Stylometry}
With the recent advances in deep learning, there exists a large body of work in the literature which employs deep neural networks for stylometry and authorship attribution. For instance,
Ge et al. used a feed forward neural network language model on an authorship attribution task. The output achieves promising results compared to the n-gram baseline \cite{ge2016authorship}. 
Bagnall et al. have employed a recurrent neural network with a shared recurrent state which outperforms other proposed methods in PAN 2015 task \cite{bagnall2016authorship}. 

Methods that particularly use CNN for stylometry application include the following.
Shrestha et al. applied CNN based on character n-gram to identify the authors of tweets. Given that each tweet is short in nature, their approach shows that a sequence of character n-grams as an  to CNN allows the architecture to capture the character-level interactions, which afterwards is aggregated
to learn higher-level patterns for modeling
the style \cite{shrestha2017convolutional}.  
Hitchler et al. propose a CNN based on pretrained embedding word vector concatenated with one hot encoding of POS tags; however, they have not shown any ablation study to report the contribution of POS tags on the final performance results \cite{hitschler2017authorship}. 
Alharthi et al. propose a book recommendation system, using an author prediction task to learn a representation which is transferable for a book recommendation process \cite{alharthi2018authorship}.






\section{The Proposed Model: Syntactic Recurrent Neural Network}\label{proposed_method}

\begin{figure*}[h]
\centering
\includegraphics[scale=0.47]{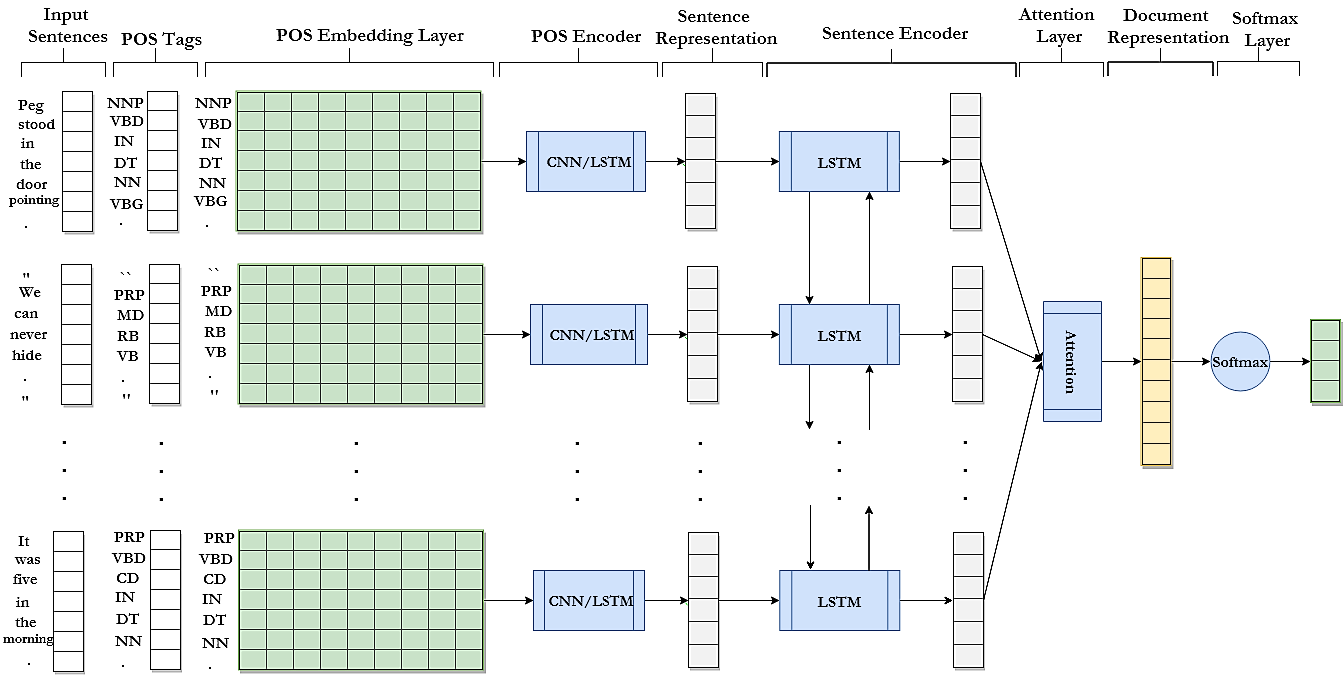}
\caption{The Overall Architecture of Syntactic Recurrent Neural Network for Style-based Text Classification}\label{posembedding}
\end{figure*}

We introduce a syntactic recurrent neural network to encode the syntactic patterns of a document in a hierarchical structure. First, we represent each sentence as a sequence of POS tags and each POS tag is embedded into a low dimensional vector and a POS encoder (which can be a CNN or LSTM) learns the syntactic representation of sentences. Subsequently, the learned sentence representations aggregate into the document representation. Moreover, we use attention mechanism to reward the sentences which contribute more to the prediction of labels. Afterwards we use a softmax classifier to compute the probability distribution over class labels. The overall architecture of the network is shown in figure \ref{posembedding}. In the following sections, we elaborate the main components of the model.

\subsection{POS Embedding}
We assume that each document is a sequence of $M$ sentences and each sentence is a sequence of $N$ words, where $M$, and $N$ are model hyperparameters and the best values are explored through the hyperparameter tuning phase (Section \ref{Tuning}). Given a sentence, we convert each word into the corresponding POS tag in the sentence and afterwards we embed each POS tag into a low dimensional vector $P_{i} \in \mathbb{R}^{d_p}$ using a trainable lookup table $ \theta_P \in \mathbb{R}^{|T|\times d_p}$, where $T$ is the set of all possible POS tags in the language. We use NLTK part-of-speech tagger \cite{bird2009natural} for the tagging purpose and use the set of $47$ POS tags\footnote{\href{https://github.com/nltk/nltk/blob/new-corpus-view/nltk/app/chunkparser\_app.py}{https://github.com/nltk/nltk/blob/new-corpus-view/nltk/app/chunkparser\_app.py}} in our model as follows.\\

\textit{\small{ T = \{ CC, CD, DT, EX, FW, IN, JJ, JJR, JJS, LS, MD, NN, NNS, NNP, NNPS, PDT, POS, PRP, PRP\$, RB, RBR, RBS, RP, SYM, TO, UH, VB, VBD, VBG, VBN, VBP, VBZ, WDT, WP, WP\$, WRB, 
\lq,\rq, 
\lq:\rq, 
\lq...\rq, 
\lq;\rq, 
\lq?\rq, 
\lq!\rq, 
\lq.\rq, 
\lq\$\rq, 
\lq(\rq, 
\lq)\rq, 
\lq `` \rq, 
\lq '' \rq
\} 
}}\\

One of the advantages of using POS tags instead of words is its low dimensional lookup table compared to the word embeddings, where the size of vocabulary in large datasets usually surpasses 50K words. On the other hand, the size of POS embedding lookup table is significantly smaller, fixed, and independent of the dataset which makes the proposed model less likely to have out-of-vocabulary words.


\subsection{POS Encoder}
POS encoder learns the syntactic representation of sentences from the output of POS embedding layer. In order to investigate the effect of short-term and long-term dependencies of POS tags in the sentence, we exploit both CNNs and LSTMs.

\subsubsection{Short-term Dependencies}
CNNs generally capture the short-term dependencies of words in the sentences which make them robust to the varying length of sentences in the documents. Lexical based CNN models have been used widely for text classification and sentiment analysis \cite{johnson2014effective, wang2012end,kim2014convolutional,collobert2011natural} and they generally outperform the conventional n-gram vector-based methods. 


Let $S_i = [P_1;P_2;...;P_N]$ be the vector representation of sentence $i$ and $W \in \mathbb{R}^{rd_p}$ be the convolutional filter with receptive field size of $r$. We apply a single layer of convolving filters with varying window sizes as the  of rectified linear unit function (relu) with a bias term b, followed by a temporal max-pooling layer which returns only the maximum value of each feature map $C^{r}_{i} \in \mathbb{R}^{N-r+1}$.
Consequently, each sentence is represented by its most important syntactic n-grams, independent of their position in the sentence.
Variable receptive field sizes $Z$ are used to compute vectors for different n-grams in parallel and they are concatenated into a final feature vector $h_i \in \mathbb{R}^{K}$ afterwards,  where $K$ is the total number of filters:\\
$$ C^{r}_{ij} = relu(W^T S_{j:j+r-1}+b), j \in [1,N-r+1] ,$$
$$ \hat{C^{r}_i} =  max\{C^{r}_i\} , $$
$$ h_i = \oplus \hat{C^{r}_i},     \forall r \in Z $$

\subsubsection{Long-term Dependencies}

Recurrent neural networks especially LSTMs are capable of capturing the long-term relations in sequences which make them more effective compared to the conventional n-gram models where increasing the length of sequences results a sparse matrix representation of documents. Lexical-based recurrent neural networks have been widely used for text classification tasks \cite{tang2015document, yang2016hierarchical}.

Let $S_i = [P_1;P_2;...;P_N]$ be the vector representation of sentence $i$. As an alternative to CNN, we use a bidirectional LSTM to encode each sentence. The forward LSTM reads the sentence $S_i$ from $P_{1}$ to $P_{N}$ and the backward LSTM reads the sentence from $P_{N}$ to $P_{1}$. The feature vector $h^{p}_{t} \in \mathbb{R}^{2d_l}$ is concatenation of the forward LSTM and the backward LSTM, where $d_l$ is the dimensionality of the hidden state. The final vector representation of sentence $i$, $h^{s}_{i} \in \mathbb{R}^{2d_l}$ is computed as unweighted sum of the learned vector representation of POS tags in the sentence. This allows us to represent a sentence by its overall syntactic pattern.
$$\overrightarrow{h^{p}_{t}} =  LSTM(P_{t}) , t \in[1, N],$$
$$ \overleftarrow{h^{p}_{t}} = LSTM(P_{t}) , t \in[N, 1],$$
$$ h^{p}_{t} = [\overrightarrow{h^{p}_{t}};\overleftarrow{h^{p}_{t}}]$$
$$h^{s}_{i} = \sum_{t \in [1,N]} h^{p}_{t}$$

\subsection{Sentence Encoder}

Sentence encoder learns the syntactic representation of a document from the sequence of sentence representations outputted from the POS encoder. We use a bidirectional LSTM To capture how sentences with different syntactic patterns are structured in a document. The outputted vector from the sentence encoder is calculated as follows.
$$\overrightarrow{h^{d}_{i}} =  LSTM(h^{s}_{i}) , i \in[1, M],$$
$$ \overleftarrow{h^{d}_{i}} = LSTM(h^{s}_{i}) , i \in[M, 1],$$
$$h^{d}_{i} = [\overrightarrow{h^{d}_{i}};\overleftarrow{h^{d}_{i}}]$$


Needless to say, not all sentences are equally informative about the authorial style of a document. Therefore, we incorporate attention mechanism to reveal the sentences that contribute more in detecting the writing style. We define a sentence level vector $u_s$ and use it to measure the importance of the sentence $i$ as follows:
$$ u_i = tanh(W_s h^{d}_{i} +b_s)$$
$$ \alpha_i=\frac{exp(u_i^T u_s)}{\sum_i exp(u_i^T u_s)} $$
$$ V = \sum _i \alpha_i h^{d}_{i}$$

Where $u_s$ is a learnable vector and is randomly initialized during the training process and $V$ is the vector representation of document which is weighted sum of vector representations of all sentences.

\begin{table*}[t!]
\begin{center}
\small
\begin{tabular}{|c||c|c||c|c||c|c|}
      \hline        & \multicolumn{2}{c||}{Train Data I} & \multicolumn{2}{c||}{Train Data II} & \multicolumn{2}{c|}{Test Data} \\  \hline
             & Word Count    & Sentence Length    & Word Count    & Sentence Length    & Word Count  & Sentence Length  \\ \hline 
Candidate 01 & 73,449        & 17                 & 76,602        & 19                 & 70,112      & 20               \\  
Candidate 02 & 180,660       & 13                 & 117,024       & 14                 & 82,317      & 13               \\  
Candidate 03 & 158,306       & 17                 & 121,301       & 19                 & 151,049     & 15               \\  
Candidate 04 & 84,080        & 14                 & 79,413        & 18                 & 93,055      & 14               \\  
Candidate 05 & 109,857       & 18                 & 141,086       & 15                 & 96,663      & 15               \\  
Candidate 06 & 61,644        & 19                 & 46,549        & 16                 & 42,808      & 16               \\  
Candidate 07 & 71,106        & 16                 & 70,563        & 18                 & 84,996      & 21               \\  
Candidate 08 & 106,024       & 18                 & 113,475       & 15                 & 94,700      & 13               \\  
Candidate 09 & 66,840        & 15                 & 41,093        & 15                 & 194,547     & 15               \\  
Candidate 10 & 86,681        & 14                 & 35,699        & 16                 & 60,998      & 16               \\  
Candidate 11 & 53,960        & 19                 & 48,037        & 13                 & 80,330      & 24               \\  
Candidate 12 & 49,543        & 25                 & 64,495        & 26                 & 50,636      & 27               \\  
Candidate 13 & 32,900        & 21                 & 153,994       & 32                 & 77,780      & 27               \\  
Candidate 14 & 89,908        & 23                 & 71,058        & 22                 & 52,633      & 35               \\  
\hline
\end{tabular}
\end{center}
\caption{\label{corpouswordstat} Corpust Statistics. }
\end{table*}

\subsection{Classification}

The learned vector representation of documents are fed into a softmax classifier to compute the probability distribution of class labels. Suppose $V_k$ is the vector representation of document $k$ learned by the attention layer. The prediction $\tilde{y_k}$ is the output of softmax layer and is computed as:
$$ \tilde{y_k} = softmax(W_{c} V_{k} + b_c)$$

Where $W_c$ , $b_c$ are learnable weight and learnable bias respectively and $\tilde{y_i}$ is a $C$ dimensional vector (C is the number of classes). We use cross-entropy loss to measure the discrepancy of predictions and true labels $y_k$. The model parameters are optimized to minimize the cross-entropy loss over all the documents in the training corpus. Hence, the regularized loss function over $N$ documents denoted by $J(\theta)$ is: 
$$ J(\theta) = - \frac{1}{N} \sum_{i=1}^{N} \sum_{k=1}^{C} y_{ik} log \tilde{y}_{ik}+ \lambda ||\theta|| $$

\section{Experimental Results}\label{experiment_eval}

\subsection{Dataset}\label{dataset}

We evaluate our proposed method on a commonly used benchmark dataset from PAN 2012 authorship attribution shared task\footnote{https://pan.webis.de/clef12/pan12-web/author-identification.html}. We chose Task I dataset which corresponds to the authorship attribution among a closed set of 14 authors. The training set comprises 28 novel-length documents (two per candidate author), ranging from 32,000 words up to about 180,000 words. The test set consists of 14 novels (one per candidate author) with the length ranging from 42,000 words up to 190,000 words.
Table \ref{corpouswordstat} reports the word count and the averaged sentence length of documents in both train and test set for each candidate author. 

In order to generate enough train/test samples, we have schematized the novels into the segments with a $M$ number of sentences (sequence length). The best value of $M$ is explored through the hyperparameter tuning phase (Section \ref{Tuning}). Accordingly, the performance measures include segment-level categorical accuracy as well as document-level categorical accuracy. In the latter, we use majority voting to label a document based on the segment-level predictions.





\subsection{Baselines}
For our baselines, we employ standard syntactic n-gram model as a syntactic approach and word n-gram model as a lexical approach. For both models, we have used Support Vector Machine (SVM) classifier with linear kernel.  Moreover, in order to compare the performance of syntactic recurrent neural network to the lexical based approaches, we fed the sequence of words to a neural network with the identical architecture. We use 300 dimensional pretrained Glove embeddings \cite{pennington2014glove} for the embedding layer in the network. In order to reduces the effect of out-of-vocabulary problem, we retain only 50,000 most frequent words.


\subsection{Hyperparameter Tuning}\label{Tuning}
In this part we examine the effect of different hyperparameters on the performance of the proposed model. All the performance metrics are the mean of segment-level accuracy (on the test set) calculated over 10 runs with 0.9/0.1 train/validation split. We use Nadam optimizer \cite{sutskever2013importance} to optimize the cross entropy loss over 30 epochs of training. 


\subsubsection{CNN for POS encoding}

\begin{figure}[h]
\includegraphics[scale=0.28]{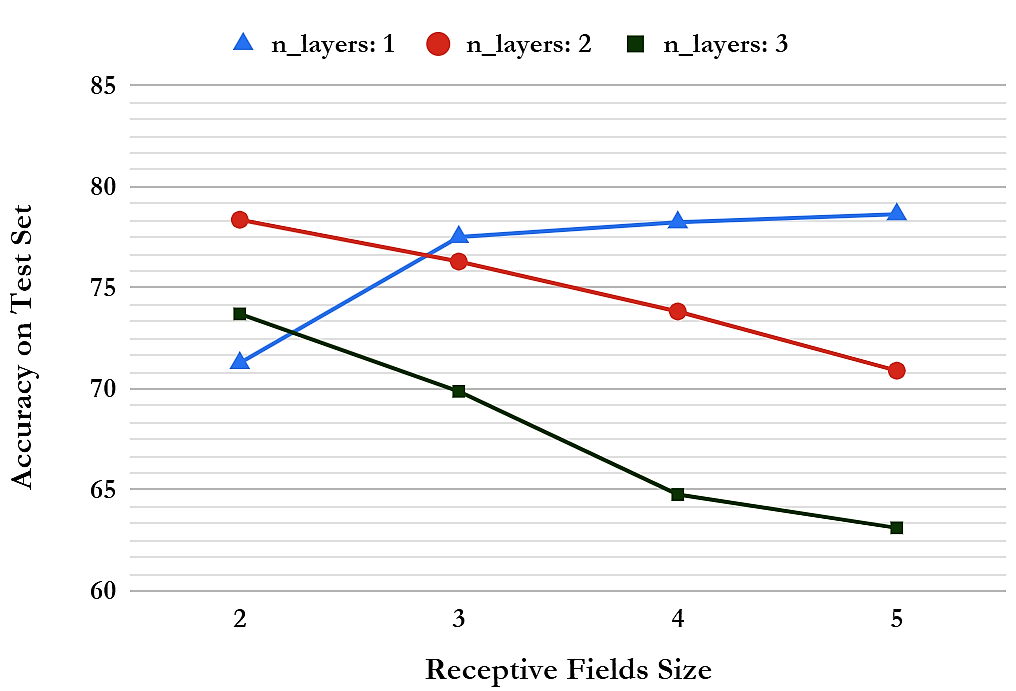}
\caption{The effect of different receptive fields sizes and number of layers (n\_layers) on the performance of syntactic recurrent neural network}\label{cnn_hyp_II}
\end{figure}

Figure \ref{cnn_hyp_II} illustrates the performance of syntactic recurrent neural network when CNN is used as POS encoder, across different receptive field sizes and number of layers while other parameters are kept constant. We observe that, increasing the number of convolutional layers generally lessens the performance. This can be due to the fact that each layer adds to the complexity of model which yields to the higher number of parameters and limited training data aggravates the performance of the model.
Moreover, in  one convolutional layer, the accuracy generally increases by increasing the size of receptive fields simply because receptive fields with the higher sizes capture longer syntactic sequences which are more informative. 

In our experiments, we also observed that having parallel convolutional layers with different receptive fields sizes improves the performance. Therefore, in the final model, we use one layer of multiple convolutional filters with the receptive filed sizes of 3 and 5.

\begin{table*}[]\centering
\begin{tabular}{|l|l|c|c|c|}
\hline
\multicolumn{2}{|c|}{\textbf{Model}} & \multicolumn{2}{c|}{\textbf{ Segment-Level Accuracy (\%)}} & \multicolumn{1}{c|}{\textbf{Document-Level Accuracy(\%)}}  \\
\cline{3-4}
\multicolumn{2}{|c|}{} &\multicolumn{1}{c|}{\textbf{Validation}} &\multicolumn{1}{c|}{\textbf{Test}} & \multicolumn{1}{|c|}{} \\
\cline{1-5} \hline
                    & Word N-grams-SVM   &90.71          &58.35              &78.57 (11/14 novels)  \\ \cline{2-5} 
\textbf{Lexical}    & CNN-LSTM          &\textbf{98.88} & 64.12             & 78.57  (11/14 novels) \\ \cline{2-5}
                    & LSTM-LSTM         &96.83          & 63.92             & 85.71  (12/14 novels)\\ \cline{2-5} \hline
                    & POS N-grams-SVM    &89.60          & 69.66             & 92.85 (13/14 novels) \\ \cline{2-5} 
\textbf{Syntactic}  & CNN-LSTM          &93.22  & \textbf{78.76}   & \textbf{100.00} (14/14 novels) \\  \cline{2-5} 
                    & LSTM-LSTM         &95.00          & 74.40        & \textbf{100.00}  (14/14 novels) \\ \hline
\end{tabular}
\caption{The performance results of models on PAN 2012 dataset for authorship attribution task.} \label{accuracy_table}
\end{table*}

\subsubsection{LSTM for POS encoding}

\begin{figure}[h]
\centering
\includegraphics[scale=0.28]{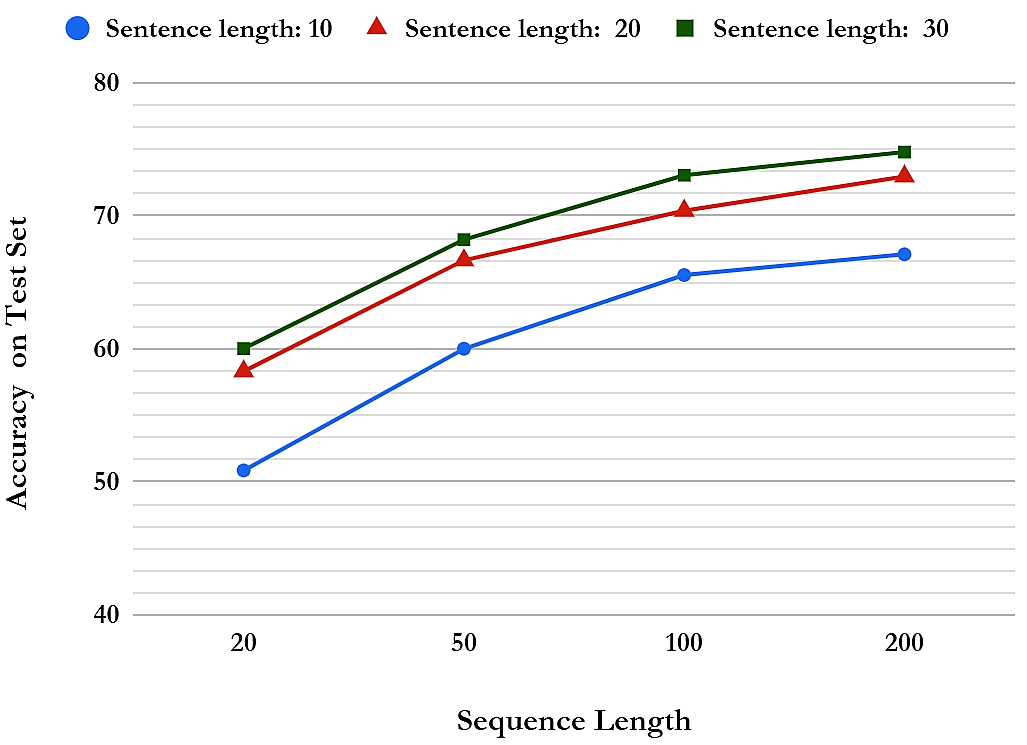}
\caption{The effect of sentence length and sequence length on the performance of syntactic recurrent neural network}\label{hnn_hyp}
\end{figure}

Figure \ref{hnn_hyp} demonstrates the accuracy of the proposed model when LSTM is employed as POS encoder, across different values of sentence length ($N$) and sequence length ($M$: the number of sentences in each segment). We observe from the figure that increasing the sequence length boosts the performance and the model achieves higher accuracy on the segments with 100 sentences (74.40) than the segments with only 20 sentences (60.02). This observation confirms that investigation of writing style in short documents is more challenging \cite{neal2017surveying}. 

As shown in the table \ref{corpouswordstat}, the average sentence length in the dataset ranges from 13 to 35. Therefore, we have examined the sentence length of 10, 20, 30, and 40 (the performance of the model is identical when the sentence length is 30 and 40, so we have not included the latter results in the figure). We observe that increasing the length of sentences to 30 words improves the performance primarily because decreasing the sentence length ignores several words in the sentence which leads to notable information loss. To sum up, syntactic neural network accepts segments as the inputs where each segment contain 100 sentences and the length of each sentence is 30.

\subsection{Results}

We report both segment-level and document-level accuracy. As mentioned before, each document (novel) has been divided into the segments of 100 sentences. Therefore, each segment in a novel has classified independently and afterwards the label of each document is calculated as the majority voting of its constituent segments. Table \ref{accuracy_table} reports the performance results of baselines and the proposed model (with both CNN and LSTM as POS encoder) on the PAN 2012 dataset. According to the segment-level accuracy, the performance of all models has dropped significantly on the test set mainly because of insufficient training data. We expect that if the models are trained on enough writing samples per author, the test results would be closer to the validation results.


Unsurprisingly, syntactic CNN-LSTM model outperforms the conventional POS n-gram model (POS N-gram-SVM) by $9.1\%$ improvement in segment-level accuracy and $7.15\%$ improvement in document-level accuracy. This is primarily because syntactic CNN-LSTM not only represents a sentence by its important syntactic n-grams but also learns how these sentences are structured in a document. On the other hand, POS N-gram-SVM model only captures the frequency of different n-grams in the document.

\subsubsection{Syntactic v.s. Lexical}

According to the table \ref{accuracy_table}, both syntactic recurrent neural networks (CNN-LSTM and LSTM-LSTM) outperform the lexical models by achieving the highest document-level accuracy ($100.00\%$).  Syntactic recurrent neural networks have correctly classified all the 14 novels in the test set while lexical LSTM-LSTM achieves the highest document-level accuracy ($85.71\%$) in the lexical models by correctly classifying 12 novels. 

In segment-level classification, syntactic recurrent neural networks outperform the lexical models in the test time with $14\%$ higher accuracy; however, the lexical models achieve higher validation accuracy. This observation may imply the lower generalization capability of lexical models compared to the syntactic models in the style-based text classification. 


\subsubsection{Short-Term v.s. Long-Term}
According to the results in table \ref{accuracy_table}, syntactic CNN-LSTM model slightly outperforms syntactic LSTM-LSTM by approximately $4\%$ in segment-level accuracy. The primary difference of two models is the way they represent a sentence. In syntactic CNN-LSTM, each sentence is represented by its important syntactic n-gram independent of their position in the sentence. However, syntactic LSTM-LSTM mainly captures the overall syntactic pattern of a sentence by summing up all the learned vector representations of POS tags in the sentence.




\subsubsection{Short Documents v.s. Long Documents}
\begin{figure}[h]
\includegraphics[scale=0.36]{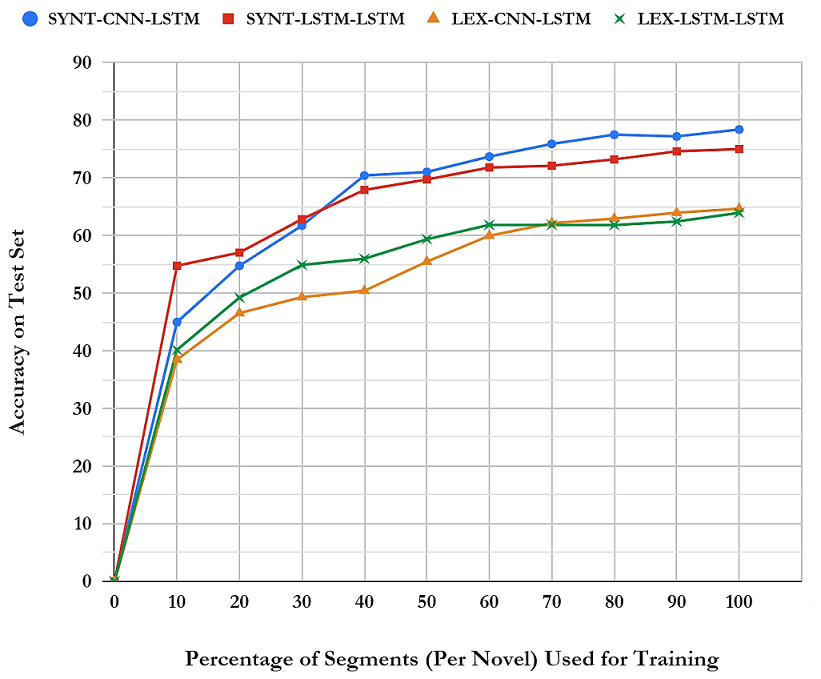}
\caption{The performance of CNN-LSTM and LSTM-LSTM models when trained on the different number of segments per document}\label{acc-perc}
\end{figure}

We have conducted a controlled study on the effect of document length on the performance of both CNN-LSTM and LSTM-LSTM models. For this purpose, we have trained each model on only specific fraction of each training document and afterwards tested the trained model on the whole test set. We keep the number of model parameters in both models approximately equal to eliminate the effect of data limitation on the training process. Figure \ref{acc-perc} demonstrates the performance results of models when trained on the first $n\%$ of segments in each document. 

\begin{figure*}[h]
\centering
\subfigure[]{\includegraphics[scale=0.35]{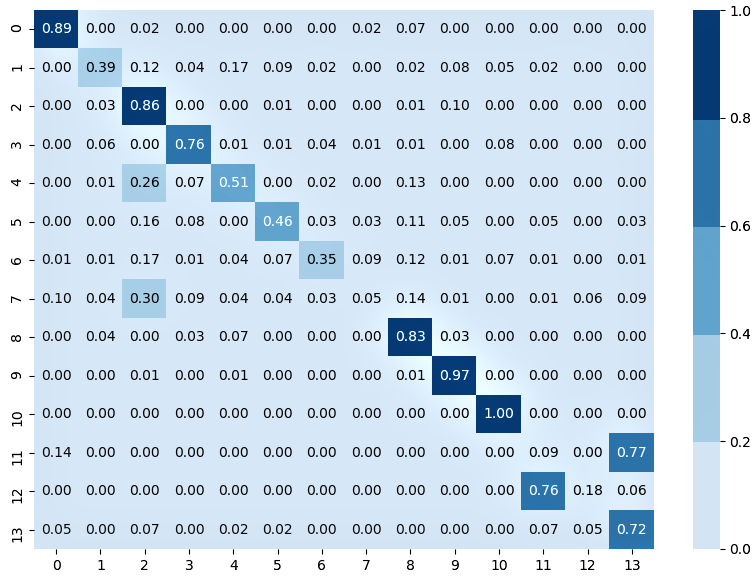}}
\subfigure[]{\includegraphics[scale=0.35]{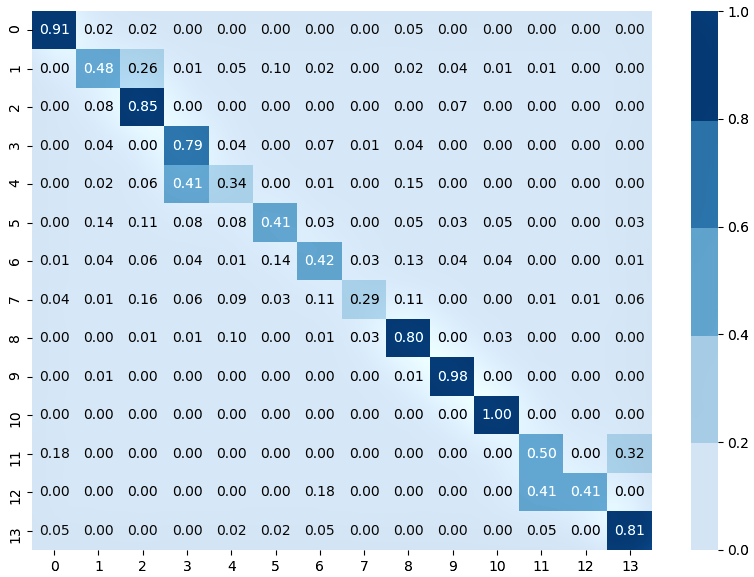}}
\subfigure[]{\includegraphics[scale=0.35]{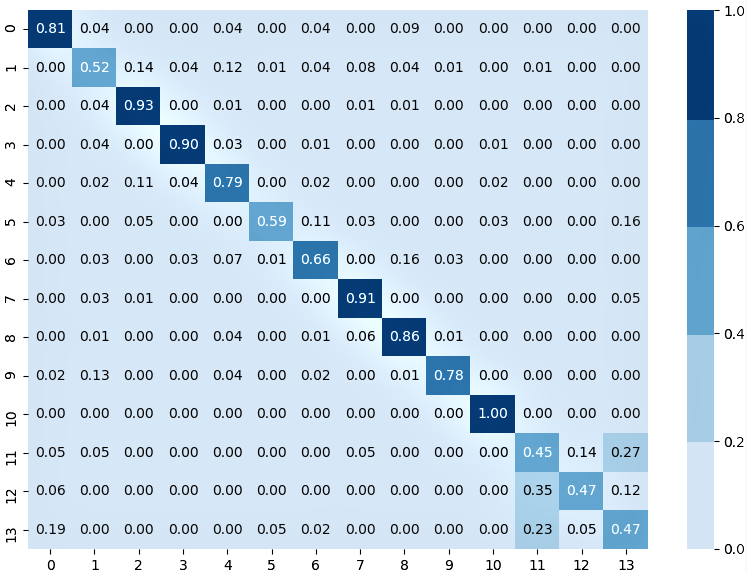}}
\subfigure[]{\includegraphics[scale=0.35]{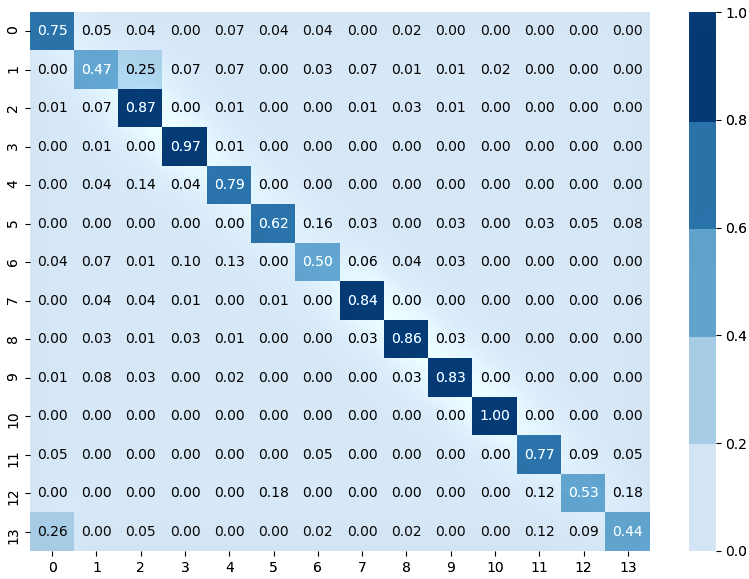}}

\caption{The confusion matrices of lexical and syntactic recurrent neural network. The labels in vertical and horizontal axis indicate class labels. (a) Lexical CNN-LSTM  model (b) Lexical LSTM-LSTM model (c) Syntactic CNN-LSTM model (d) Syntactic LSTM-LSTM model}  \label{confmatrix}
\end{figure*}

We observe that when the smaller portion of segments ($<30\%$) are used for training, LSTM-LSTM models achieve higher test accuracy than CNN-LSTM models in both syntactic and lexical settings. On the other hand, CNN-LSTM models slightly outperform LSTM-LSTM models when the number of segments used for training in each document increases. On the other words, LSTM-LSTM models appear to be quicker in capturing authorial writing style than CNN-LSTM models which this property makes them a preferred potential model when investigating authorial writing style in a dataset of short documents.

\subsubsection{Class-wise Performance}
Figure \ref{confmatrix} illustrates the segment-level recall for each class label for both lexical (a and b) and syntactic recurrent neural networks (c and d). Cell [i,j] reports the fraction of segments in document written by author i where attributed to author j. In lexical networks, LSTM-LSTM have lower miss-classification rate (2 incorrectly classified documents) than CNN (3 incorrectly classified documents).  Syntactic CNN-LSTM and LSTM-LSTM achieve the highest recall and correctly classify all the 14 documents in the test set. Both lexical models have relatively low recall in class labels 1,4,7,11 and 12  while both syntactic models show low recall in class label 13. Moreover, both lexical models as well as syntactic CNN-LSTM show lower recall for class label 11 and 12; however, syntactic LSTM-LSTM shows a higher recall in these classes.


\section{Conclusion and Future Work}\label{conclusions}
In this paper, we introduced a syntactic recurrent neural network in order to encode the syntactic patterns of documents in a hierarchical structure and afterwards used the learned syntactic representation of document for style-based text classification. We investigated both long-term and short-term dependencies of part-of-speech (POS) tags in sentences. According to our experimental results on PAN 2012 dataset, syntactic recurrent neural networks outperform lexical based networks by $14\%$ in terms of segment-level accuracy. Moreover, we observed that LSTM-based POS encoders are quicker in capturing the authorial writing style than CNN-based POS encoders which this property makes them a preferable model when investigating authorial writing style in a dataset of short documents.

\newpage
\bibliography{references.bib}
\bibliographystyle{acl_natbib}

\end{document}